\documentclass[letterpaper]{article} 
\usepackage[submission]{aaai23} 
\usepackage{times}  
\usepackage{helvet}  
\usepackage{courier}  
\usepackage[hyphens]{url}  
\usepackage{graphicx} 
\usepackage{natbib}  
\usepackage{caption} 
\usepackage{algorithm}
\usepackage{algorithmic}
\usepackage{amssymb}
\usepackage{booktabs}
\usepackage{xcolor}

\usepackage{newfloat}
\usepackage{listings}
\DeclareCaptionStyle{ruled}{labelfont=normalfont,labelsep=colon,strut=off} 
\floatstyle{ruled}
\newfloat{listing}{tb}{lst}{}
\floatname{listing}{Listing}
\usepackage[inline]{enumitem}
\usepackage{amsmath}

\def\eqref#1{equation~\ref{#1}}

\def\1{\bm{1}}

\DeclareMathAlphabet{\mathsfit}{\encodingdefault}{\sfdefault}{m}{sl}
\SetMathAlphabet{\mathsfit}{bold}{\encodingdefault}{\sfdefault}{bx}{n}

\newcommand{\valstd}[2]{$#1$ & $#2$}
\newcommand{\valstdb}[2]{$\mathbf{#1}$ & $#2$}

\title{An End-to-End Reinforcement Learning Approach for\\ Job-Shop Scheduling Problems Based on Constraint Programming}
\author{
    Pierre Tassel\textsuperscript{\rm 1},
    Martin Gebser\textsuperscript{\rm 1,\rm 2}, 
    Konstantin Schekotihin \textsuperscript{\rm 1}
}
\affiliations{
    \textsuperscript{\rm 1}University of Klagenfurt,\\
    \textsuperscript{\rm 2}Graz University of Technology\\
    \{pierre.tassel, martin.gebser, konstantin.schekotihin\}@aau.at
}

\begin{document}

\maketitle


\begin{abstract}
  Constraint Programming (CP) is a declarative programming paradigm that allows for modeling and solving combinatorial optimization problems, such as the Job-Shop Scheduling Problem (JSSP).
  While CP solvers manage to find optimal or near-optimal solutions for small instances, they do not scale well to large ones, i.e., they require long computation times or yield low-quality solutions.
  Therefore, real-world scheduling applications often resort to fast, handcrafted, priority-based dispatching heuristics to find a good initial solution and then refine it using optimization methods.

  This paper proposes a novel end-to-end approach to solving scheduling problems by means of CP and Reinforcement Learning (RL).
  In contrast to previous RL methods, tailored for a given problem by including procedural simulation algorithms, complex feature engineering, or handcrafted reward functions, our neural-network architecture and training algorithm merely require a generic CP encoding of some scheduling problem along with a set of small instances.   
  Our approach leverages existing CP solvers to train an agent learning a Priority Dispatching Rule (PDR) that generalizes well to large instances, even from separate datasets.
  We evaluate our method on seven JSSP datasets from the literature, showing its ability to find higher-quality solutions for very large instances than obtained by static PDRs and by a
  CP solver within the same time limit.
\end{abstract}


\section{Introduction}\label{sec:introduction}
    Scheduling problems are a class of combinatorial optimization problems widely used in many domains, such as manufacturing, logistics, or healthcare.
    They are characterized by a set of \emph{jobs} associated with a list of \emph{operations} that must be executed on some of the available \emph{machines}.
    The goal is to find a schedule that minimizes an objective function, such as the makespan (i.e., the total completion time), flow-time (i.e., average job processing time), or total tardiness.
    Scheduling problems can be modeled using Constraint Programming (CP), which is a declarative programming paradigm allowing for efficient modeling of combinatorial optimization problems.
    Having this layer of abstraction has several advantages.
    Namely, it enables \emph{(i)} a compact representation of the problem, which is easier to implement and maintain than a procedural approach, and \emph{(ii)} application of domain-independent high-performance solving algorithms \cite{rossiHandbookConstraintProgramming2006}.
    %
    However, CP approaches scale poorly to large instances. 
    While small instances can be solved in a matter of seconds, the computation time increases significantly with the instance size since most scheduling problems are known to be NP-complete or NP-hard \cite{kanGeneralFlowshopJobshop1976,DBLP:journals/mor/GareyJS76,SOTSKOV1995237}.
    This is particularly problematic for applications where scheduling must be performed in (near) real-time or where instances evolve dynamically, e.g., in the re-scheduling context.

    As a consequence, real-world scheduling applications often rely on fast, but handcrafted heuristics, like Priority Dispatching Rules (PDRs), to find some good initial solution that can further be improved by more time-consuming meta-heuristics or exact methods such as CP or Mixed Integer Programming (MIP).
    While PDRs can find initial solutions quickly, their design requires much domain-specific knowledge and, still, they often yield limited or inconsistent performance depending on the instance at hand.
    Due to the widespread adoption of PDRs, a large body of literature has been devoted to developing such heuristics for scheduling problems \cite{DBLP:journals/ior/PanwalkarI77}.
    As a recent trend, a variety of automatic approaches to learning PDR heuristics using Reinforcement Learning (RL) have been proposed. 
    
    During training, an RL agent interacts with the environment and learns from the reward it receives. However, when applied to scheduling problems, the learning is complicated by the fact that a reward is only received at the end of an episode when the schedule is complete. Thus, the environment provides only a poor feedback signal to the agent, requiring manual shaping of rewards for a particular problem definition \cite{DBLP:reference/ml/Wiewiora10, DBLP:conf/socs/OrenRLRTFCD21}.
    %
    %
    Hence, while RL approaches have shown promising results, e.g., \cite{ingimundardottirDiscoveringDispatchingRules2018, linSmartManufacturingScheduling2019a, zhangLearningDispatchJob2020e, tasselReinforcementLearningEnvironment2021c}, they rely on custom, procedural simulation algorithms tailored for a given problem, sophisticated feature engineering, large disjunctive graph representations, and handcrafted reward functions.
    Since these approaches select one operation to dispatch at a time, they also often suffer from performance issues as the environment triggers a forward pass in the neural network for each operation.
    This is particularly problematic for instances with a large number of operations to dispatch, in which case the performance advantage of a learned PDR heuristic can be dramatically reduced.

    \paragraph*{Contributions.}
    In this paper, we propose a novel RL approach that interacts with a CP solver directly to find a solution for a scheduling problem without requiring any custom observation features or handcrafted reward functions. Our contributions can be summarized as follows:\footnote{Live demo: \url{https://pierretassel-jobshopcprl.hf.space}\\
    Command to install the environment: \emph{pip install job-shop-cp-env}\\
    Source code available at: \url{https://github.com/ingambe/End2End-Job-Shop-Scheduling-CP}%
    }%
    \begin{description}[nosep,leftmargin=5pt,labelindent=-5pt]
    
      \item \textbf{Constraint Programming Environment.} We propose a novel RL environment comprising a CP model representing the scheduling problem, similar to the one used by CP solvers to find solutions.
    The raw CP model variables represent the environment's state.
    The agent interacts with the environment by selecting a variable to fix, and the environment propagates the constraints to the other variables.
    Our environment uses lazy instantiation and allocates a variable-size set of operations per iteration, allowing us to model large scheduling problems and improve solving performance by reducing the number of forward passes.
    \item \textbf{Learning Algorithm.} We present an algorithm that does not require any custom feature engineering or handcrafted reward function to train the agent.
    Our approach leverages a CP solver to provide feedback and search trajectories guiding the learning process.
    As a result, the training on 
    small instances generalizes well to larger ones, even from separate datasets.
    \item \textbf{Neural Network Architecture.} We introduce a neural network architecture able to extract information from the CP model and efficiently select a variable to instantiate.
    Our architecture is composed of encoder transformer layers applied at several levels to extract a representation of the current state of the problem, followed by a point-wise multi-layer perceptron to obtain the priority of each operation.
    \item \textbf{Solving Method.} To generate solutions to a scheduling problem, we use a simulated-annealing-like meta-heuristic to select the operation to allocate in the environment based on multiple actors in parallel.
    In particular, when an agent has a high
    associated temperature, it is likely to 
    explore alternative operations,
    while prioritized operations are more greedily preferred otherwise.
    Hence, using our agent for dispatching leads to high-quality solutions quickly, even for large instances. Empirically, our method establishes an unmatched state of the art for learned PDR heuristics to solve the Job-Shop Scheduling Problem (JSSP).
    \end{description}

\section{Preliminaries}\label{sec:preliminaries}

\paragraph{Job-Shop Scheduling Problem.} JSSP is a classic optimization problem \cite{thompsonIndustrialScheduling1963}. Each JSSP instance comprises sets of jobs $\mathcal{J}$ and machines $\mathcal{M}$.
Each job $j \in \mathcal{J}$ has an associated set of operation $\mathcal{O}_{j}$, and each operation $o \in \mathcal{O}_{j}$ has a unique associated machine $m \in \mathcal{M}$ and processing time $p_{o} \in \mathbb{N^+}$.
The goal is to find an assignment of starting times to all operations that minimizes a given objective function, usually the makespan. In addition, this assignment must satisfy the following constraints: jobs cannot be preempted, a machine can only process one operation at a time, and an operation cannot start before its predecessor is completed.
JSSP is NP-complete \cite{DBLP:journals/mor/GareyJS76}, and no tractable algorithms for solving it are known. 

\paragraph{Constraint Programming} CP has been widely adopted in the industry to solve scheduling problems of various sizes \cite{dacolIndustrialsizeJobShop2022}.
Each CP model comprises a set of \emph{variables} with their respective \emph{domains} and \emph{constraints} that define the relationships between them. Every time a variable is assigned, the CP solver uses \emph{constraint propagation} to filter domains of other variables, eliminating values from their domains that can never be part of a 
solution (as they violate the constraints). This process maintains \emph{consistency} of the model and allows for efficiently \emph{pruning} the search space. A \emph{solution} to a CP model is an assignment of values to variables such that all constraints are satisfied.
While early CP solvers were based on simple backtracking search that did not scale well, modern solvers rely on nogood learning, meta-heuristics, and relaxation techniques that iteratively improve an initial solution \cite{laborieSelfAdaptingLargeNeighborhood,laborieTemporalLinearRelaxation2016,vilimFailureDirectedSearchConstraintBased2015}.
Also in the scheduling context, starting with a good initial solution helps the solving process to converge to a high-quality solution faster \cite{kovacsUtilizingConstraintOptimization2021a}.

\paragraph{Reinforcement learning.} RL is a machine learning approach, which trains an agent to solve a task by interacting with its environment \cite{DBLP:books/lib/SuttonB98}. The environment, described by a \emph{Markov Decision Process} (MDP), provides the agent with an \emph{observation} that describes the current state of the environment, based on which the agent selects an \emph{action} to apply to the environment. This process is repeated until the agent reaches a terminal state. Periodically, the agent receives a \emph{reward} from the environment, indicating how expedient the performed action was, where the reward is called \emph{continuous} if the agent receives a reward after each action, and \emph{episodic} otherwise. The goal of the agent is to learn a \emph{policy} $\pi$ mapping each observation to an action that maximizes the \emph{cumulative reward}.

\section{Related Work}\label{sec:related-work}

Recently, the idea of applying RL to solve combinatorial optimization problems has gained a lot of attention \cite{bengioMachineLearningCombinatorial2021b}.
Successful applications of RL to problems like Traveling Salesperson \cite{belloNeuralCombinatorialOptimization2022}, Vehicle Routing \cite{koolAttentionLearnSolve2022}, and Boolean Satisfiability \cite{yolcuLearningLocalSearch2019} fueled the interest of the operations research community, especially for large-scale, industrial instances.

Similarly, various approaches to train RL agents on scheduling problems have been proposed.
One of them \cite{linSmartManufacturingScheduling2019a} involves selecting a PDR per machine from a pool of predefined candidates. 
\citet{linSmartManufacturingScheduling2019a} employ a Deep Q-Network (DQN), 
trained on a collection of manually designed features for customer orders and system states.
Their method assumes a fixed number of jobs and machines, and an underlying neural network is fine-tuned to each instance, resulting in a longer overall runtime than taken by CP solvers.
However, 
benchmark results show that the proposed approach can outperform handcrafted PDRs.
Likewise, \citet{teppanGeneticAlgorithmsCreating2020} came to 
comparable conclusions with a complementary method 
using Genetic Algorithms.


The idea to exploit imitation learning from a MIP solver has been explored by~\citet{ingimundardottirDiscoveringDispatchingRules2018}. 
They aim to learn a composition of pre-defined PDR candidates by learning from a dataset of optimal solutions generated by a MIP solver.
However, because of the JPPS complexity, only small instances of $10$ jobs $\times$ $10$ machines are considered. Their approach also requires defining handcrafted features for the observation space provided by a procedural environment.
Moreover, policies learned by pure imitation learning tend to generalize poorly outside the training set observed by the agent.

Given that a JSSP instance can be represented as a disjunctive graph, \citet{zhangLearningDispatchJob2020e} train a Graph Neural Network to allocate operations using a custom procedural environment and a specifically designed continuous reward function.
Evaluations show that their approach can learn a better PDR than the handcrafted ones from the literature. However, solving large instances with thousands of operations is problematic, since the induced 
graph representations contain a large number of nodes and edges corresponding to all possible orders of the operations in an instance. 

Finally, \citet{tasselReinforcementLearningEnvironment2021c} suggest an optimized RL environment for JSSP.
They designed handcrafted observation features for the jobs and formulated a continuous reward function mimicking the makespan objective.
The environment contains several optimizations, such as prioritizing jobs that are not at their final operation over those that are and enforcing a compressed solution by preventing operations that would certainly lead to a sub-optimal state.
Although 
promising results were obtained by ten-minute training of an agent with a simple fully-connected neural network using the Proximal Policy Optimization (PPO) algorithm, 
their approach could not outperform a state-of-the-art CP solver.
Moreover, the neural network is instance-size specific and requires re-training from scratch for each new instance with a different number of jobs.

\section{Method}\label{sec:method}


In this paper, we propose a method that, in contrast to previous work, relies on a CP model for JSSP, which provides the training environment of an RL agent 
and its neural network. 
\subsection{Constraint Programming Model}\label{subsec:cp-model}

\begin{algorithm}[tb]
  \caption{CP Model for JSSP}
  \label{alg:cp_model}
  \textbf{Input}:
  \begin{itemize}[leftmargin=2pt]
    \item[] \begin{tabular}{l l@{~}r}
      $p_{o} \in \mathbb{N^+}$ & $\forall o \in \mathcal{O}_j$ & (processing time of operation $o$)\\
    \end{tabular}
    \item[] \begin{tabular}{l l@{~}r}
      $s_{o_1, o_2} \in \{0, 1\}$ & $\forall o_1, o_2 \in \mathcal{O}_j$ & (successor operations)\\ 
    \end{tabular}
    \item[] \begin{tabular}{l@{~~~}l@{~}r}
      $w^m_{o} \in \{0, 1\}$ & $\forall o \in \mathcal{O}_j, m \in \mathcal{M}$ & (machine $m$ operations)\\ 
    \end{tabular}
  \end{itemize}
  \textbf{Variables}: 
  \begin{itemize}[leftmargin=2pt]
    \item[] \begin{tabular}{lll}
      $\textrm{IntervalVariable(}k_{o}\textrm{)}$ & $\forall o \in \mathcal{O}_j$ &\\
    \end{tabular}
    \item[] \begin{tabular}{lll}
      with & $k_{o}.length, k_{o}.start, k_{o}.end \in \mathbb{N^+}$ &\\
    \end{tabular}
  \end{itemize}
  \textbf{Constraints}: 
  \begin{itemize}[leftmargin=2pt]
    \item[] \begin{tabular}{lll}
      $k_{o}.length = p_{o}$ & $\forall o \in \mathcal{O}_j$ & \\
    \end{tabular}
    \item[] \begin{tabular}{lll}
      $s_{o_1, o_2} = 1 \Rightarrow k_{o_1}.end \leqslant k_{o_2}.start$ & $\forall o_1, o_2 \in \mathcal{O}_j$ &\\
    \end{tabular}
    \item[] \begin{tabular}{lll}
      $\textrm{NoOverlap(} \{ o \mid w^m_{o} = 1 \} \textrm{)}$ & $\forall m \in \mathcal{M}$ &\\
    \end{tabular}
  \end{itemize}
  \textbf{Objective}: 
  \begin{itemize}[leftmargin=2pt]
    \item[] \begin{tabular}{lll}
      \textit{minimize} & $\textrm{max(}k_{o}.end\textrm{)}$ & $\forall o \in \bigcup_{j \in \mathcal{J}}\mathcal{O}_j$ \\
    \end{tabular}
  \end{itemize}
\end{algorithm}

Our environment is based on a classic CP model for JSSP as described in Algorithm~\ref{alg:cp_model}.
While compact, this model is very expressive and powerful. The global \emph{NoOverlap} constraint is a key component of the model ensuring that no two operations are executed simultaneously on the same machine. 
In addition, the precedence constraint ensures that the operations of each job are executed in their predefined order. 

\subsection{Environment}\label{subsec:env}

At each decision point, the current time $t$ is defined by the minimum lower bound $k_o^{lb}.end$ among all interval variables, i.e., $\forall o \in \bigcup_{j \in \mathcal{J}}\mathcal{O}_j \; k_o^{lb}.end \leq k_o.end$.
To interact with the CP model, our environment fixes the start time of the current operation of a selected job to its lower bound. This is done by adding a constraint to the model and asking for propagation.
While usually all operations are passed to the model at once, we lazy load the intervals using an $n$-operations horizon per job to improve performance.
For instance, given $n=10$, only the first $10$ operations of each job are passed to the model at the beginning of the schedule.
Each time an operation is allocated, we add the next operation of the job and its associated constraints to the model.
This allows us to keep the model small and avoids propagating the whole schedule at each step, thus improving the propagation time.

\paragraph{State Space $\mathcal{S}$.}\label{par:state_space}
For each job, the environment provides the previous interval variable, the interval variable of the current operation, and the next $3$ interval variables to allocate.
Considering more upcoming operations helps the agent to achieve slightly better performance, but it also increases the state space size and the computational cost of the environment.
As the upper bound on the number of interval variables per job 
is fixed, the state representation's space complexity is $\mathsf{O}(|\mathcal{J}|)$ and thus linear with respect to the number of jobs.
This makes our method more scalable to large instances than
other approaches relying on the disjunctive graph representation of JSSP, which encodes the whole schedule as a graph and yields the higher space complexity $\mathsf{O}(|\mathcal{J}| \times |\mathcal{M}|)$. 

\paragraph{Action Space $\mathcal{A}$.}\label{par:action_space}
The action space is represented by a Boolean vector comprising \emph{(i)} all jobs that can be allocated at the current time~$t$, and \emph{(ii)} a \emph{No-Op} action enabling the agent to skip the current time step without allocating any operation.
That is, the action space is a vector of size (up to) $|\mathcal{J}| + 1$.
Whenever \emph{No-Op} is selected, the environment sets $t=k_o'.end$ such that $k_o^{lb}.end < k_o'.end$ and there is no $k_o''$ for which $k_o^{lb}.end < k_o''.end < k_o'.end$.

In addition, the agent can reduce the number of allocation steps by providing an ordered vector of actions to the environment.
The environment applies actions in the given order until no more allocations can be done at time~$t$.
In this way, a vector of actions decreases the number of state transitions and network inferences for improving the performance in terms of wall-clock time.
For training purposes, the vector of actions allocated by the environment is also returned to the agent.
A similar technique has been successfully applied in a gradient-free RL context for the Resource-Constrained Scheduling Problem \cite{tasselReinforcementLearningDispatching2022b}.

\paragraph{Reward Function $r_t$.}\label{par:reward_function}
Only the terminal state returns the objective function value of the CP model (i.e., the makespan) as a reward.
This means that the reward function is episodic.

\subsection{CP Solution Space}\label{subsec:cp-solution-space}

\begin{figure}[b!]
  \centering
   \includegraphics[width=\linewidth]{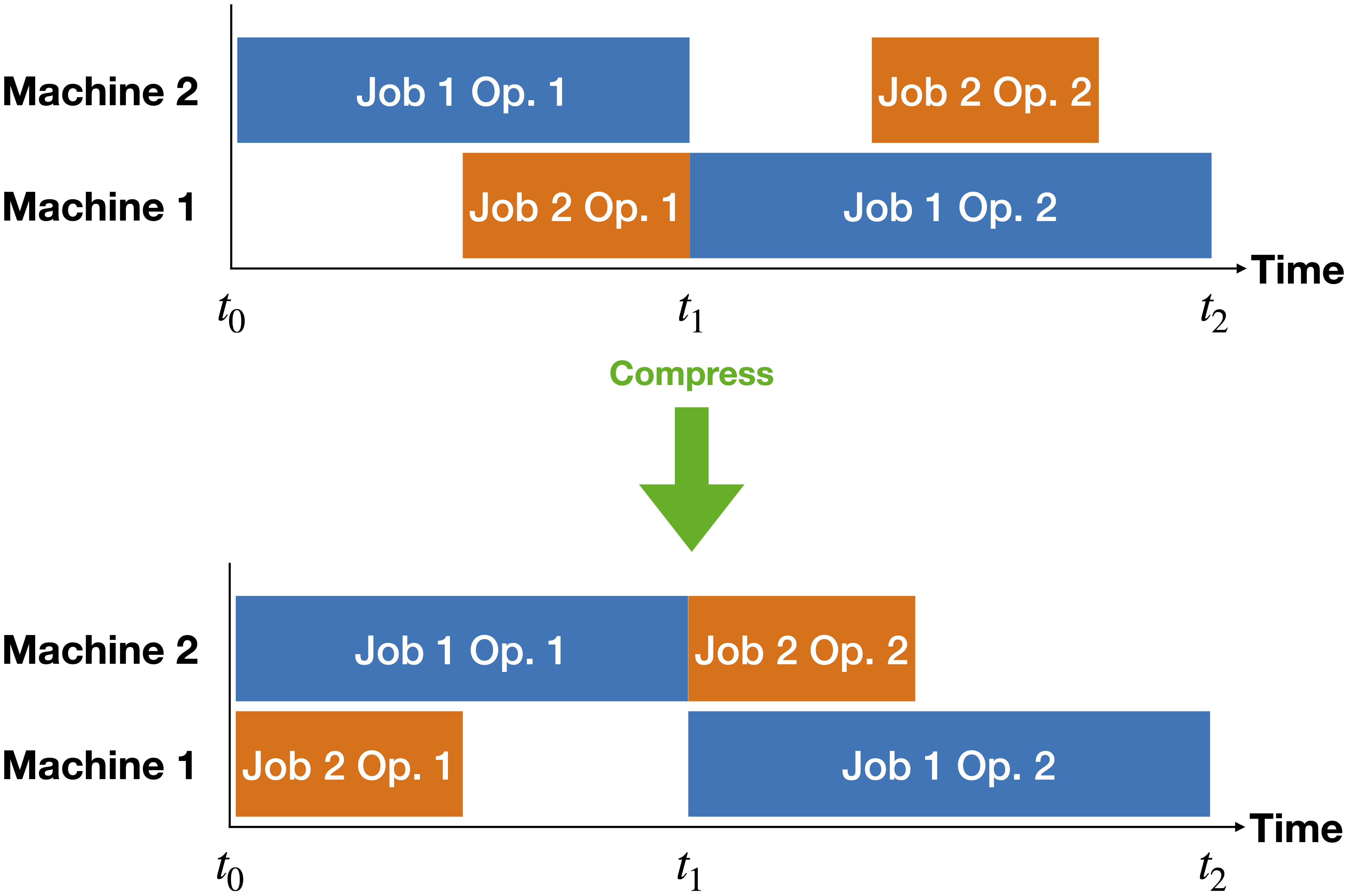}
   \caption{Example of solution compression. The resulting compressed solution has the same objective value as the original solution. However, the starting time of each operation is assigned to its lower bound.}
   \label{fig:compression_example}
\end{figure}

While a CP solver can generate all solutions found by the environment, the inverse is not the case.
The environment fixes the starting time of each operation greedily to its lower bound, whereas CP has more degrees of freedom and can assign starting times in the interval $[\text{lower\_bound}, \text{upper\_bound}]$.
Therefore, obtained solutions might be uncompressed, i.e., 
the starting time of some operation may not be assigned to its lower bound.
For example, the solution 
shown in Figure~\ref{fig:compression_example} 
has an uncompressed allocation of the operations $o_1$ and $o_2$ of the job $j_2$, as $o_2$ does not start at the completion time of $o_1$.
The goal of compression is to enforce the starting time of each operation to be assigned to its lower bound, which streamlines the solution's structure without affecting its objective value.

Compression can be done in polynomial time by setting the starting time of each operation to the lower bound of the interval variable after the initial propagation.
Algorithm~\ref{alg:compress_cp_model} describes the CP model used to compress a solution.
On top of the original constraints of the JSSP model, it adds constraints enforcing the starting time of each operation to be lower or equal to the starting time of the uncompressed solution.
Another constraint preserves the order of operations on each machine.

As our environment relies on a CP model, it can, at any time, take advantage of state-of-the-art CP solvers to generate a solution.
The solution can then be compressed using the additional constraints described in Algorithm~\ref{alg:compress_cp_model} to generate a solution compatible with the environment.
Therefore, partial solutions that complete the initial solution generated by the actor can be produced.
We exploit this property of our environment during the training phase to generate example solutions and compare the actor's decisions with solutions generated by the CP solver.

\begin{algorithm}[bt]
  \caption{CP Model to Compress JSSP Solution}
  \label{alg:compress_cp_model}
  \textbf{Given an initial solution $D$ for a JSSP instance} \\
  \textbf{Input}:
  \begin{itemize}[leftmargin=2pt]
    \item[] \begin{tabular}{l l@{~}r}
      $p_{o} \in \mathbb{N^+}$ & $\forall o \in \mathcal{O}_j$ & (processing time of operation $o$)\\
    \end{tabular}
    \item[] \begin{tabular}{l l@{~}r}
      $s_{o_1, o_2} \in \{0, 1\}$ & $\forall o_1, o_2 \in \mathcal{O}_j$ & (successor operations)\\ 
    \end{tabular}
    \item[] \begin{tabular}{l@{~~~}l@{~}r}
      $w^m_{o} \in \{0, 1\}$ & $\forall o \in \mathcal{O}_j, m \in \mathcal{M}$ & (machine $m$ operations)\\ 
    \end{tabular}
    \item[] \begin{tabular}{l l r}
      $u_{o} \in \mathbb{N^+}$ & $\forall o \in \mathcal{O}_j$ & (starting time in $D$)\\
    \end{tabular}
  \end{itemize}
  \textbf{Variables}: 
  \begin{itemize}[leftmargin=2pt]
    \item[] \begin{tabular}{lll}
      $\textrm{IntervalVariable(}k_{o}\textrm{)}$ & $\forall o \in \mathcal{O}_j$ &\\
    \end{tabular}
    \item[] \begin{tabular}{lll}
      with & $k_{o}.length, k_{o}.start, k_{o}.end \in \mathbb{N^+}$ &\\
    \end{tabular}
  \end{itemize}
  \textbf{Constraints}: 
  \begin{itemize}[leftmargin=2pt]
    \item[] \begin{tabular}{lll}
      $k_{o}.start \leq u_{o}$ & $\forall o \in \mathcal{O}_j$ & \\
    \end{tabular}
    \item[] \begin{tabular}{lll}
      $k_{o}.length = p_{o}$ & $\forall o \in \mathcal{O}_j$ & \\
    \end{tabular}
    \item[] \begin{tabular}{lll}
      $s_{o_1, o_2} = 1 \Rightarrow k_{o_1}.end \leqslant k_{o_2}.start$ & $\forall o_1, o_2 \in \mathcal{O}_j$ &\\
    \end{tabular}
    \item[] \begin{tabular}{lll}
      $\textrm{NoOverlap(} \{ o \mid w^m_{o} = 1 \} \textrm{)}$ &  $\forall m \in \mathcal{M}$ 
       &\\
    \end{tabular}
    \item[] \begin{tabular}{lll}
      $w^m_{o_1} + w^m_{o_2} = 2 \wedge u_{o_1} < u_{o_2}\Rightarrow k_{o_1}.end \leqslant k_{o_2}.start$ &\\
      \multicolumn{2}{@{}r}{$\forall o_1, o_2 \in \bigcup_{j \in \mathcal{J}} \mathcal{O}_j, m \in \mathcal{M}$}\\
    \end{tabular}
  \end{itemize}
  \textbf{Objective}: 
  \begin{itemize}[leftmargin=2pt]
    \item[] \begin{tabular}{lll}
      \textit{minimize} & $\textrm{sum(}k_{o}.start\textrm{)}$ & $\forall o \in  \bigcup_{j \in \mathcal{J}} \mathcal{O}_j$ \\
    \end{tabular}
  \end{itemize}
\end{algorithm}

\begin{figure*}[tb]
  \centering
   \includegraphics[width=0.839\textwidth]{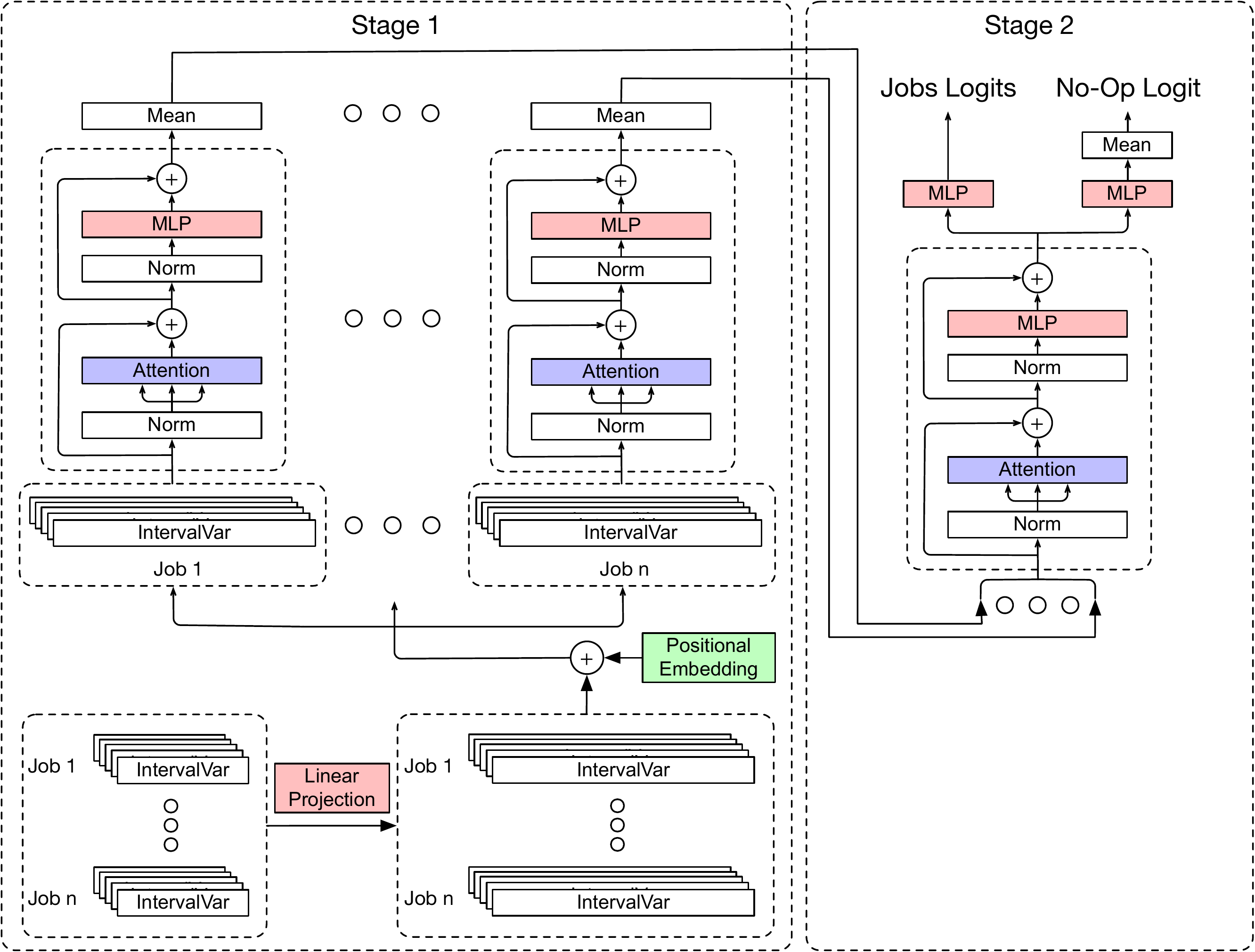}
   \caption{Policy network architecture. The network is composed of $2$ stages. The first stage extracts a job's representation out of the interval variables of the job. The jobs' representations are then fed to the second stage, which outputs a priority score for each job and the \emph{No-Op} action.}   \label{fig:nn_arch}
\end{figure*}

\subsection{Neural Network Architecture}\label{subsec:neural-network-arch}

The agent policy is a function $\pi_\theta(\mathcal{A}\mid s_t)$ that maps the current state of the environment to a probability distribution over the set of actions $\mathcal{A}$.
The policy is parameterized by parameters $\theta$ of a neural network, whose structure is illustrated in Figure~\ref{fig:nn_arch}.
As discussed in the state space definition above, for each job, the environment provides interval variables representing the previous, current, and $3$ next operations of each job.
The neural network is composed of two stages: the first stage generates a per-job representation of the state, while the second stage determines the action distribution using the previously generated job representations.

In the first stage, the neural network self-learns a representation for the source and sink interval variables representing the start and end operations of a job. This approach is common for neural networks, encoding a sequence by its start and end elements with specific tokens, and all elements in-between using their relative position with respect to these tokens.
Each interval variable is encoded as a quadruple  $(f, lb, l, ct)$, where $f=1$ if the environment has already assigned a value to the interval variable and $f=0$ otherwise; $lb,l \in \mathbb{N}^+$ correspond to the lower bound on the starting time and the processing time of an operation, respectively; and $ct = 1$ if $lb = t$, and $ct =0$ otherwise.
This $4$-dimensional space is projected to an $8$-dimensional space using a fully connected layer in order to improve the performance of the transformer architecture. 
Then, the network adds the resulting vector with $8$ components to an absolute and fixed positional encoding, which represents interval variable positions by their relative distance to the latest predecessor operation already allocated by the environment.
For example,
if the environment has allocated the sub-sequence $(o_1, o_2)$ of operations $\mathcal{O}_j= (o_1, o_2, \dots, o_n)$ for a job $j$, positions of the interval variables $k_{o_3}, \dots, k_{o_n}$ will be encoded relative to the variable $k_{o_2}$.
This positional encoding enforces the precedence constraint between interval variables with respect to the currently allocated operations of a job.
All extended interval variable representations are then passed to a \emph{Transformer Encoder Layer} \cite{vaswaniAttentionAllYou2017a} to generate a representation for each job.

In the second stage, each job's representation is passed to another \emph{Transformer Encoder Layer}, and then to a \emph{Multi-Layer Perceptron} (MLP) with one hidden layer of dimension $32$ and hyperbolic tangent activation in-between to generate the \emph{logits} of the job allocation action.
The logits vector is concatenated to the average of the output of another MLP of a similar dimension that generates the logits vector of the \emph{No-Op} action.
The resulting logits vector, of dimension $|\mathcal{J}| + 1$, is finally passed to a softmax layer to determine the probability distribution over the set of actions.

\section{Training and Evaluation Algorithms}\label{sec:training-algorithm}

The training algorithm can be described as a hybrid between \emph{Imitation Learning} and \emph{Policy Gradient} methods.
While pure imitation learning methods learn from expert demonstrations, policy gradient methods learn from the agent's own experiences.
Pure imitation learning approaches tend to generalize poorly, while policy gradient methods require a continuous reward function to learn from.
The hybrid method we propose combines the best of both worlds by learning from both expert demonstrations and the actor's experiences with feedback received from the expert.
Due to the architecture of the environment, having access to an expert is straightforward since a CP solver can directly interact with the learner and provide sample solutions.

\subsection{Partial Expert Demonstrations and Feedback}\label{subsec:generating-expert-demos-feedback}

Algorithm~\ref{alg:expert_demonstration} describes the process used to generate partial expert demonstrations and feedback from an initial actor's trajectory.
At each iteration, each actor $a \in A$ collects (in parallel) one episode of experiences by interacting with the environment.
Afterward, the algorithm randomly selects a number $j$ between $0$ and the smallest actors' episode length and keeps the first $j$ steps of each actor's episode as a partial solution.
Next, the complete solution of an actor is forwarded to a CP solver, which tries to improve it.
The obtained solution is then compressed using the CP model presented in Algorithm~\ref{alg:compress_cp_model} to generate a solution compatible with the environment.
Because the search for (partial) solutions is intractable, we limit the CP solver's runtime to $t$ seconds.
This timeout criterion, initially set to $60$ seconds, is increased by $1$ second after each iteration, allowing the CP solver to find better solutions as the training progresses.

\begin{algorithm}[tb]
  \caption{Partial Expert Demonstration Generation Using Actors and CP Solver Trajectories}
  \label{alg:expert_demonstration}
  \textbf{Parameter}: $t$: maximum admitted CP solver runtime.\\
  \textbf{Output}: For actor $a \in A$:
  \begin{itemize}
    \item[] $\mathcal{B}_a$: An initial solution of length $j$ generated by the actor.
    \item[] $\mathcal{C}_a$: Partial solution generated by the actor starting from the initial solution $\mathcal{B}_a$.
    \item[] $\mathcal{D}_a$: Partial solution generated by the CP solver in $t$ seconds starting from the initial solution $\mathcal{B}_a$.
  \end{itemize}
  \begin{algorithmic}
    \FOR{actor $a \in A$}
      \STATE Complete one episode.      
    \ENDFOR

    \STATE Sample $j$ uniformly from $[0, \text{min\_episode\_length}]$
    \FOR{actor $a \in A$}
      \STATE $\mathcal{B}_a \gets$ trajectory of actor $w$ up to the $j$-th step.
      \STATE $\mathcal{C}_a \gets$ partial solution by the actor completing $\mathcal{B}_a$.
      \STATE $\mathcal{D}_a \gets$ partial solution by the CP solver completing $\mathcal{B}_a$, warm-starting from $\mathcal{C}_a$ as initial assignment.
    \ENDFOR
  \end{algorithmic}
\end{algorithm}

\subsection{Training Using Feedback and Examples From CP}\label{subsec:CP-feedback}

The generated trajectories are provided to Algorithm~\ref{alg:feedback_actor}, which first evaluates the quality of the actor's trajectories. The quality of such solutions is measured as a ratio of the solution makespan generated by the CP solver over the solution makespan generated by the actor. The smaller the ratio, the worse the quality of the actor's solution. Algorithm~\ref{alg:feedback_actor} uses the ratio to penalize actions made by the actor and reward actions made by the CP solver. Since the actor's solution is passed to the CP solver as a search starting point, the CP solver's solution is always at least as good as the actor's. That is, if a CP solver could not improve the solution, the training algorithm penalizes no actions. 

\begin{algorithm}[tb]
  \caption{Training Algorithm Using CP Solver's Feedback and Examples}
  \label{alg:feedback_actor}
  \textbf{Input}: $\theta$: neural network parameters.\\
  For actor $a\in A$: 
  \begin{itemize}
    \item[] $\mathcal{C}_a$: Partial solution generated by the actor.
    \item[] $\mathcal{D}_a$: Partial solution generated by the CP solver.
    \item[] $i \in \mathcal{I}_a$: Improvement ratio of the solution generated by the CP solver over the solution generated by the actor.
    \item[] $\mathcal{R}_a$: $-i$ improvement for trajectory generated by the actor, $i$ otherwise.
  \end{itemize}
  \textbf{Parameter}: $K$: Maximum number of training iterations. \\
  $\beta$: Maximum KL divergence allowed between SGD updates.
  \begin{algorithmic}
    \STATE Min-Max scaled advantage of $\mathcal{R}$: $\mathcal{N.R} \gets$  $\frac{\mathcal{R} - \text{min(}\mathcal{R}\text{)}}{\text{max(}\mathcal{R}\text{)} - \text{min(}\mathcal{R}\text{)}}$.
    \FOR{iteration $i=1,\dots,K$}
    \STATE Sample batch $b$ of experiences:
    \begin{itemize}
      \item[] $obs$: Observation of the environment.
      \item[] $action$: Action either taken by the actor or CP solver.
      \item[] $\mathcal{N.R}_b$: Normalized improvement.
    \end{itemize}
    \STATE $L = \max \left(r(\theta), \operatorname{clip}\left(r(\theta), 1-\epsilon, 1+\epsilon\right) \right)$ \\
    $\text{where }r(\theta)= -\mathcal{N.R}_b * \frac{\pi_{\theta}\left(action \mid obs\right)}{\pi_{\theta_{\text {old }}}\left(action \mid obs\right)}$.
    \STATE Back-propagate loss $L$.
    \IF{$\text{KL}[\pi_{\theta} \mid \pi_{\theta_{\text {old }}}] > \beta$}
      \STATE \textbf{break}
    \ENDIF
    \ENDFOR
  \end{algorithmic}
\end{algorithm}

Technically, the actor's improvements are \emph{Min-Max Scaled} to be in the $[0,1]$ range, and the training is done using a loss function with a clipped surrogate objective to prevent the policy from changing too abruptly between iterations \cite{schulmanProximalPolicyOptimization2017c}.
The advantage function used to estimate the policy gradient is the negative \emph{Min-Max Scaled} improvement.

\subsection{Improving Initial Solutions}\label{subsec:improving_initial_solutions}

While Algorithm~\ref{alg:feedback_actor} is capable of improving the agent's decisions taken to complete the initial solution using partial expert demonstrations, it cannot improve the initial solution relative to which the expert demonstrations are collected.
To address this issue, we propose penalizing/rewarding the agent's initial solution collected by the actor based on the quality of the solution generated by the CP solver.
Intuitively, if the initial solution is of low quality, the solution found by the CP solver will be worse than with an initial solution of better quality.
Algorithm~\ref{alg:feedback_initial} is used to improve the initial solutions passed to the CP solver.
The normalized CP solver's objective function value (i.e., 
the makespan) is used as the advantage function to estimate the policy gradient.
Therefore, the 
initial solutions leading to the best result of the CP solver in an iteration will be reinforced, whereas the worst ones will be penalized.

\begin{algorithm}[tb]
  \caption{Training Algorithm Improving Initial Solutions}
  \label{alg:feedback_initial}
  \textbf{Input}: $\theta$: neural network parameters.\\
  For actor $a\in A$:
  \begin{itemize}
    \item[] $\mathcal{C}_a$: Partial solution generated by the actor.
    \item[] $\mathcal{R}_a$: Makespan of the solution generated by the CP solver starting from $\mathcal{C}_a$.
  \end{itemize}
  \textbf{Parameter}: $K$: Maximum number of training iterations. \\
  $\beta$: Maximum KL divergence allowed between SGD updates. \\
  $\epsilon$: Surrogate clipping coefficient.
  \begin{algorithmic}
    \STATE Normalize advantage of $\mathcal{R}$: $\mathcal{N.R} \gets$  $\frac{\mathcal{R} - \text{mean(}\mathcal{R}\text{)}}{\text{std(}\mathcal{R}\text{)}}$.
    \FOR{iteration $i=1,\dots,K$}
    \STATE Sample batch $b$ of experiences:
    \begin{itemize}
      \item[] $obs$: Observation of the environment.
      \item[] $action$: Action taken by the actor.
      \item[] $\mathcal{N.R}_b$: Normalized improvement.
    \end{itemize}
    \STATE $L = \max \left(r(\theta), \operatorname{clip}\left(r(\theta), 1-\epsilon, 1+\epsilon\right) \right)$ \\
    $\text{where }r(\theta)= \mathcal{N.R}_b * \frac{\pi_{\theta}\left(action \mid obs\right)}{\pi_{\theta_{\text {old }}}\left(action \mid obs\right)}$.
    \STATE Back-propagate loss $L$.
    \IF{$\text{KL}[\pi_{\theta} \mid \pi_{\theta_{\text {old }}}] > \beta$}
      \STATE \textbf{break}
    \ENDIF
    \ENDFOR
  \end{algorithmic}
\end{algorithm}

\subsection{Solution Generation}\label{subsec:solution_generation}

The algorithm used to generate solutions from the previously trained neural network is described in Algorithm~\ref{alg:eval_actor}.
This algorithm uses different temperatures on each parallel actor to generate solutions in the neighborhood of the trained policy.
Actors having a high temperature will be more likely to select a different action than the top action predicted by the neural network.
In contrast, low-temperate actors will sample more greedily from the top actions predicted by the neural network.
As the actors are run in parallel, the algorithm can efficiently explore separate parts of the solution space, generating better solutions than greedy sampling.

\begin{algorithm}[tb]
  \caption{Data Collection}
  \label{alg:eval_actor}
  \textbf{Input}: $\theta$: neural network parameters. 

  \begin{algorithmic}
    \FOR{each actor $a\in A$} 
      \STATE Temperature parameter $\mathcal{T}_{a} \gets (1.5 \times \frac{a}{|A|}) + 0.5$.
      \STATE Initialize environment $\mathcal{E}_a$.
      \STATE Empty solution $\mathcal{S}_a$.
    \ENDFOR
    \WHILE{not all actors have terminated}
      \FOR{actor $a\in A$ 
           in parallel}
        \IF{actor $a$ has not terminated}
          \STATE $obs \gets$ observation of the environment $\mathcal{E}_a$.
          \STATE $logits \gets$ neural network $\theta$ applied to $obs$.
          \STATE $probabilities \gets$ softmax($\frac{logits}{\mathcal{T}_a}$). 
          \STATE $action \gets$ sample from $probabilities$.
          \STATE $obs, done \gets$ step($\mathcal{E}_a$, $action$).
          \IF{$done$}
            \STATE $\mathcal{S}_a \gets$ solution generated by the actor.
          \ENDIF
        \ENDIF
      \ENDFOR
    \ENDWHILE
    \RETURN $\min_{a\in A}
             \text{makespan}(\mathcal{S}_a)$.
  \end{algorithmic}
\end{algorithm}

\section{Experiments}\label{sec:experiments}

We evaluate our method on seven 
JSSP benchmark sets, including $284$ instances in total.
These datasets cover a wide range of difficulty and size, from small instances with $36$ operations to large instances with $100,000$ operations.

\subsection{Experimental Setup}\label{sec:experimental_setup}

\begin{table*}[ht]
  \centering
\resizebox{\textwidth}{!}{
\begin{tabular}{@{} r @{~~} l @{~~} r@{$\displaystyle \,\pm\,$}l @{~~} r@{$\displaystyle \,\pm\,$}l @{~~} r@{$\displaystyle \,\pm\,$}l @{~~} r@{$\displaystyle \,\pm\,$}l @{~~} r@{$\displaystyle \,\pm\,$}l @{}}
\toprule
\textbf{Dataset} & & \multicolumn{2}{c}{\textbf{Ours}}  & \multicolumn{2}{c}{\textbf{Choco}} 
& \multicolumn{2}{c}{\textbf{FIFO}}    & \multicolumn{2}{c}{\textbf{SPT}} & \multicolumn{2}{c@{}}{\textbf{MTWR}} \\
\midrule 
\textbf{\citeauthor{taillardBenchmarksBasicScheduling1993a}} & Makespan              & \valstdb{2\,670.26}{78.84}      & \valstd{3\,045.95}{131.62}                       & \valstd{3\,165.69}{162.98}     & \valstd{3\,128.77}{131.94}   & \valstd{3\,086.18}{127.61}  \\
                & Runtime (s)               & \valstd{17.98}{0.18}        & \valstd{17.98}{0.18}                        & \valstdb{1.32}{0.01}             & \valstd{1.35}{0.07}            & \valstd{1.36}{0.10}  \\ 
\addlinespace 
\textbf{\citeauthor{demirkolBenchmarksShopScheduling1998a}} & Makespan                 & \valstdb{5\,701.53}{909.13}   & \valstd{6\,292.19}{820.93}   & \valstd{6\,397.31}{623.03}  & \valstd{6\,481.71}{868.65}   & \valstd{6\,275.99}{748.11}  \\
                                                          & Runtime (s)              & \valstd{15.12}{0.53}             & \valstd{15.12}{0.53}    & \valstd{0.79}{0.07}          & \valstd{0.78}{0.04}            & \valstdb{0.77}{0.00}  \\ 
\addlinespace 
\textbf{\citeauthor{lawrenceResouceConstrainedProject1984}} & Makespan              & \valstdb{1\,197.77}{66.61}    & \valstd{1\,253.02}{67.49}     & \valstd{1\,432.97}{91.68}  & \valstd{1\,411.15}{98.60}                     & \valstd{1\,331.50}{109.81}  \\
                                                  & Runtime (s)              & \valstd{5.71}{0.04}          & \valstd{5.71}{0.04}       & \valstdb{0.10}{0.00}            & \valstd{0.10}{0.00}                          & \valstd{0.10}{0.00}  \\ 
\addlinespace 
\textbf{\citeauthor{applegateComputationalStudyJobShop1991a}}                                 & Makespan              & \valstd{1\,056.81}{224.64}   & \valstdb{1\,027.10}{226.60}  & \valstd{1\,173.40}{254.00}  & \valstd{1\,183.60}{267.97}         & \valstd{1\,138.30}{227.66}  \\
                                                  & Runtime (s)              & \valstd{4.80}{0.04}         & \valstd{4.80}{0.04}     & \valstd{0.04}{0.00}       & \valstdb{0.04}{0.00}                 & \valstd{0.11}{0.22}  \\ 
\addlinespace 
\textbf{\citeauthor{storerNewSearchSpaces1992}}                                     & Makespan              & \valstdb{2\,398.38}{203.03}   & \valstd{2\,610.27}{201.70}  & \valstd{2\,581.77}{155.83}   & \valstd{2\,728.37}{227.39}  & \valstd{2\,532.73}{153.12}  \\
                                                  & Runtime (s)              & \valstd{9.72}{0.11}       & \valstd{9.72}{0.11}     & \valstdb{0.38}{0.00}         & \valstd{0.38}{0.00}     & \valstd{0.38}{0.00}  \\ 
\addlinespace 
\textbf{\citeauthor{yamadaGeneticAlgorithmApplicable1992}}                                     & Makespan              & \valstdb{1\,068.80}{59.76}   & \valstd{1\,116.75}{53.09}  & \valstd{1\,234.25}{116.56}  & \valstd{1\,220.25}{74.50}    & \valstd{1\,233.00}{82.60}  \\
                                                  & Runtime (s)            & \valstd{9.62}{0.02}         & \valstd{9.62}{0.02}    & \valstdb{0.29}{0.00}           & \valstd{0.29}{0.00}            & \valstd{0.29}{0.00}  \\ 
\addlinespace 
\textbf{\citeauthor{dacolIndustrialsizeJobShop2022}}                                      & Makespan              & \valstdb{147\,178.82}{1\,823.08}   & \valstd{\infty}{\infty}       & \valstd{152\,894.44}{2\,633.71}  & \valstd{150\,044.50}{2\,203.50}   & \valstd{149\,463.80}{2\,447.76}  \\
                                                  & Runtime (s)            & \valstd{316.14}{0.96}            & \valstd{316.14}{0.96}    & \valstdb{70.48}{0.52}             & \valstd{74.23}{0.78}            & \valstd{74.71}{0.82}  \\ 
\midrule
\textbf{Average} & Makespan                         & \valstdb{28\,675.83}{480.73}                  & \valstd{\infty}{\infty}              & \valstd{30\,070.92}{576.83}        & \valstd{29\,591.46}{553.22}    & \valstd{29\,391.23}{556.67}        \\
\bottomrule
\end{tabular}
}
\caption{
  For each dataset, the average makespan and runtime in seconds are reported (lower is better) along with their respective standard deviations.
  In total, our approach outperforms static PDRs and the CP solver Choco 
  running for the same time as our approach.
  For the \citeauthor{dacolIndustrialsizeJobShop2022} dataset, Choco 
  could not find a solution for any of the instances within the given time limit.%
}
\label{tbl:aggregated_results}
\end{table*}

The literature proposes hundreds of handcrafted, static PDRs for JSSP with various features and performances so that we cannot compare them exhaustively.
Hence, we selected the three most performant and dominating PDRs based on the survey by \citet{selsComparisonPriorityRules2012}.
These three PDRs are the First In First Out (\emph{FIFO}), Shortest Processing Time (\emph{SPT}), and Most Total Work Remaining (\emph{MTWR}) heuristics.
Additionally, we compare the 
open-source Choco CP solver \cite{prudhommeChocosolverJavaLibrary2022}, running for the same time 
as our approach.
Since our approach involves randomness for sampling the actor's actions, we report average metrics over $10$ runs with different random seeds.

\subsection{Training Settings and Hyperparameters}\label{subsec:training-setting}

We ran the training algorithm on $4$ instances simultaneously by applying $24$ actors to each instance, resulting in a parallel training of $96$ actors in total.
This multi-instance training is crucial to the algorithm's success. 
It targets the agent to learn a generic PDR heuristic that generalizes rather than a policy that is tailored to some specific instance.
While theoretically any combination of $4$ instances could be used, for the training algorithm to perform stable, these instances need to have approximately the same number of operations to allocate.
Otherwise, some instances would contribute more observations during the training, and thus the training process would be biased toward this instance.

We carefully selected the $4$ training instances to achieve a balance between acquiring high-quality CP traces within a reasonable timeframe and investigating diverse scheduling scenarios. 
Our experimentation led us to choose Taillard's instances with $30 \text{ jobs} \times 15 \text{ machines}$ as the ideal compromise between training difficulty and efficiency. 
We note that Taillard's instances with $30 \text{ jobs} \times 20 \text{ machines}$ provided similar outcomes, yet took longer to solve and generate acceptable expert demonstrations using CP.
We trained our agent for $100$ epochs,
each comprising
$20$ training iterations of $20$ mini-batches.
Finally, our experiments yielded $n=10$ and $3$ as empirically advantageous parameters for lazy instantiation or the number of next operations per job to
include in the state representation, respectively.

\subsection{Results}\label{sec:results}

In our evaluation, the agent trained on $30 \text{ jobs} \times 15 \text{ machines}$ Taillard's instances was applied to 
instances from seven popular JSSP benchmark sets.
The results of these experiments are summarized in Table~\ref{tbl:aggregated_results}.
For each dataset, we aggregate the instances and report the average makespan as well as the runtime in seconds.
The PDR heuristic learned by our method outperforms the compared static PDRs on every dataset. In particular, on the \textbf{\citeauthor{taillardBenchmarksBasicScheduling1993a}} dataset, we improve over MTWR, which is the best performing static PDR, by about $13\%$ in terms of the absolute makespan of solutions.
On the very large instances of the \textbf{\citeauthor{dacolIndustrialsizeJobShop2022}} dataset, our approach outperforms MTWR by just $1.5\%$, but the makespan is an absolute metric whose (unknown) optimum is greater than zero, 
so that the relative improvement with respect to optimal solutions is significantly higher.

In terms of runtime, due to neural network communication and forward pass computation, our method is slower than static PDRs,
especially on small instances.
On large instances (\textbf{\citeauthor{dacolIndustrialsizeJobShop2022}}), the gap narrows because the static PDRs always select a single action, while our agent provides an ordered vector of actions per time step.
The latter approach reduces the number of allocation steps, which apparently compensates the neural network overhead to some extent on large instances.

When comparing our approach to the CP solver Choco, 
running for the same time as our agent, we outperform the solver on all datasets except the one by \textbf{\citeauthor{applegateComputationalStudyJobShop1991a}}, which comprises very small instances with $10 \text{ jobs} \times 10 \text{ machines}$ only. As a consequence, finding solutions, even optimal ones, is relatively easy for the state-of-the-art CP solver Choco.
However, our approach consistently outperforms Choco on the other datasets.
In case of large instances, a scenario in which the use of PDRs is common practice, Choco fails to find any solution within the given time limit.

We refrain from elaborating results per size of the instances in each dataset,
but invite the reader to inspect the supplementary material for respective details. The supplement also compares the literature solutions (where available) of the disjunctive graph-based RL approach by \citet{zhangLearningDispatchJob2020e}.
Note that \citeauthor{zhangLearningDispatchJob2020e}'s method involves re-training the agent for each instance size, while our agent is only trained once on a small set of reasonably sized instances.

\section{Conclusions and Future Work}\label{sec:conclusion}

This paper presents an end-to-end Deep RL approach for solving the Job-Shop Scheduling Problem.
We provide a novel way to set up the RL environment based on a generic CP model of JSSP for state updates.
Our environment lazy loads the variables of the CP model, enabling a fast propagation of constraints, even for large instances.
We developed a size-agnostic, efficient neural network architecture capable of extracting features from the raw variables of the CP model, thus eliminating the need for custom observation designs.
We also present a novel training algorithm, leveraging the CP nature of the environment by using a CP solver to 
generate expert feedback and trajectories.
Our training method does not require any custom, continuous reward function, and it is capable of learning a PDR heuristic from one dataset that generalizes well to other, unseen datasets.
Extensive experiments on seven benchmark sets from the literature show that our approach outperforms static PDRs and the CP solver Choco 
within the same time limit, thus establishing an unmatched state of the art for learned PDR heuristics to solve JSSP.

In future work, we plan to harness our approach for solving instances extracted from real-world applications.
Since real-world instances are continuous, with repetitive patterns, RL approaches are expected to learn PDR heuristics adapting to the scheduling problem's underlying distribution.
This is not the case with datasets from the literature, where the distribution of instances is uniform.
Also, we aim to improve the efficiency of our method and to adopt it to other problems, like the Resource-Constrained Scheduling Problem.

\section*{Acknowledgments}
This work was 
funded by
KWF project 28472,
cms electronics GmbH,
FunderMax GmbH,
Hirsch Armbänder GmbH,
incubed IT GmbH,
Infineon Technologies Austria AG,
Isovolta AG,
Kostwein Holding GmbH, and
Privatstiftung Kärntner Sparkasse.

\newpage
\bibliography{aaai23}

\end{document}